\def\IncludeSupportingInformation{}
\documentclass[12pt,a4paper]{article}
\usepackage[T1]{fontenc}
\usepackage[utf8]{inputenc}
\usepackage{mathptmx}
\usepackage[margin=25mm]{geometry}
\usepackage{amsmath,amssymb,graphicx,booktabs,multirow}
\usepackage[table]{xcolor}
\usepackage{placeins,caption,setspace,tikz}
\usepackage[american]{babel}
\usepackage{csquotes}
\usepackage[backend=biber,style=numeric-comp,sorting=none,natbib=true,
  giveninits=true,maxbibnames=6,minbibnames=3,maxcitenames=2,
  doi=true,url=true,isbn=false,date=year,urldate=long]{biblatex}
\DeclareNameAlias{author}{given-family}
\DeclareFieldFormat[article,inproceedings,misc,online]{title}{\mkbibquote{#1}}
\DeclareFieldFormat{journaltitle}{\mkbibemph{#1}}
\DeclareFieldFormat{booktitle}{\mkbibemph{#1}}
\DeclareFieldFormat{volume}{#1}
\DeclareFieldFormat{number}{#1}
\DeclareFieldFormat{pages}{#1}
\DeclareFieldFormat{doi}{\url{https://doi.org/#1}}
\renewbibmacro*{in:}{}
\newbibmacro*{admd:year}{\iffieldundef{year}{\printtext{n.d.}}{\printfield{year}}}
\newbibmacro*{admd:links}{%
  \iffieldundef{doi}
    {\iffieldundef{eprint}{\usebibmacro{url+urldate}}{\usebibmacro{eprint}}}
    {\printfield{doi}}}
\DeclareBibliographyDriver{article}{%
  \usebibmacro{bibindex}\usebibmacro{begentry}%
  \printnames{author}\setunit{\addcomma\space}%
  \printfield{title}\setunit{\addcomma\space}%
  \printfield{journaltitle}\setunit{\space}\printfield{volume}%
  \iffieldundef{number}{}{\setunit{\addcomma\space}\printtext{no.\space}\printfield{number}}%
  \setunit{\space}\printtext[parens]{\usebibmacro{admd:year}}%
  \iffieldundef{pages}{}{\setunit{\addcolon\space}\printfield{pages}}%
  \iffieldundef{eid}{}{\setunit{\addcolon\space}\printfield{eid}}%
  \newunit\newblock\usebibmacro{admd:links}\usebibmacro{finentry}}
\DeclareBibliographyDriver{inproceedings}{%
  \usebibmacro{bibindex}\usebibmacro{begentry}%
  \printnames{author}\setunit{\addcomma\space}%
  \printfield{title}\setunit{\addcomma\space}%
  \printtext{in\space}\printfield{booktitle}%
  \iffieldundef{volume}{}{\setunit{\addcomma\space}\printtext{vol.\space}\printfield{volume}}%
  \setunit{\space}\printtext[parens]{\usebibmacro{admd:year}}%
  \iffieldundef{pages}{}{\setunit{\addcolon\space}\printfield{pages}}%
  \iffieldundef{eid}{}{\setunit{\addcomma\space}\printtext{Article\space}\printfield{eid}}%
  \newunit\newblock\usebibmacro{admd:links}\usebibmacro{finentry}}
\DeclareBibliographyDriver{misc}{%
  \usebibmacro{bibindex}\usebibmacro{begentry}%
  \printnames{author}\setunit{\addcomma\space}%
  \printfield{title}\setunit{\space}%
  \printtext[parens]{\usebibmacro{admd:year}}%
  \newunit\printfield{howpublished}%
  \newunit\newblock\usebibmacro{admd:links}%
  \newunit\printfield{note}\usebibmacro{finentry}}
\DeclareBibliographyAlias{online}{misc}

\usepackage[hidelinks]{hyperref}
\definecolor{bestresult}{RGB}{210,240,215}
\definecolor{worstresult}{RGB}{248,215,215}
\definecolor{review}{RGB}{160,70,0}

\begin{document}
\begingroup\singlespacing
{\Large\bfseries Synthetic Image Oversampling for Rare-Class Representation Learning in Deep Facial Phenotyping\par}
\vspace{8pt}
{\small\textbf{Miles Liao, Zhanliang Wang, Hongzhuo Chen, Kai Wang*}

M. Liao\\
Hopkins Authentic Research Program for Students, Hopkins School, New Haven, CT 06515, USA

M. Liao, Z. Wang, H. Chen, K. Wang\\
Raymond G. Perelman Center for Cellular and Molecular Therapeutics, Children's Hospital of Philadelphia, Philadelphia, PA 19104, USA

Z. Wang, H. Chen\\
Department of Mathematics, University of Pennsylvania, Philadelphia, PA 19104, USA

K. Wang\\
Department of Pathology and Laboratory Medicine, Perelman School of Medicine, University of Pennsylvania, Philadelphia, PA 19104, USA

*Correspondence: \href{mailto:wangk@chop.edu}{wangk@chop.edu}
\par}
\section*{Abstract}
Limited and unevenly distributed patient photographs constrain deep facial phenotyping for rare genetic disorders. Synthetic augmentation could supplement these data, but its value depends on recognition of real patients and the distribution of benefits across disorders. We evaluate disease-specific synthetic oversampling using 1,847 real facial images from ten disorders, with 1,289 training images and a fixed patient-disjoint evaluation partition of 558 images. Four nested augmentation levels are assessed with GestaltMatcher-Arc and GestaltMML across three training seeds per condition. Mean overall Top-1 accuracy increases from 89.19\% to a maximum of 90.20\% for GestaltMatcher-Arc and from 68.82\% to 74.31\% for GestaltMML. The Arc augmentation level with the highest overall accuracy differs from that with the highest mean per-class accuracy. At the largest augmentation level, GestaltMML improves mean accuracy for seven disorders but reduces it for three. Exploratory embedding analysis associates fewer real training images with larger gains in class separation, although geometric changes do not consistently track recognition performance. These findings support evaluating synthetic augmentation through both aggregate and disease-level outcomes, with representation geometry providing complementary evidence. Checkpoint selection and performance reporting use the same evaluation partition, and differences in image--text pairing limit attribution of the multimodal gains solely to synthetic image quantity. Independent test cohorts remain necessary to establish generalizability and clinical utility.

\par\endgroup
\clearpage
\section{Introduction}

Rare genetic disorders collectively affect a substantial population, yet individual syndromes may be encountered infrequently
\citep{nguengang2020estimating}. For disorders with characteristic craniofacial features, facial morphology can provide information for syndrome recognition and patient matching. Deep facial phenotyping uses patient photographs to support these tasks
\citep{gurovich2019identifying,hsieh2022gestaltmatcher}, but its development is constrained by the availability of training images. Collections are often uneven across disorders, leaving some classes with few examples of disease-associated facial variation
\citep{liu2019large,cui2019class,hustinx2023improving}. For a sparsely represented syndrome, the observed images may cover only a narrow portion of its phenotypic variation, making it difficult to learn features that generalize to other patients. The challenge is therefore to learn from limited observations of individual disorders as well as from unequal class frequencies. Acquiring more photographs is difficult for rare conditions, motivating approaches that supplement disease-specific training examples while retaining real patients as the target of recognition.

Existing systems address different aspects of this problem. DeepGestalt and GestaltMatcher learn from facial images, PEDIA incorporates image analysis into exome prioritization, and GestaltMML combines facial and clinical information
\citep{gurovich2019identifying,hsieh2022gestaltmatcher,hsieh2019pedia,wu2024gestaltmml}. Class weighting, margin-based losses, decoupled training, and logit adjustment can reduce the influence of class imbalance
\citep{cui2019class,cao2019learning,kang2019decoupling,menon2021longtaillearninglogitadjustment,Lin_2017_ICCV}. However, increasing a sparse class's loss weight still relies on the same observed examples and cannot supply additional observations of facial variation. Generative augmentation offers a complementary strategy, with applications in medical imaging
\citep{goodfellow2014generativeadversarialnetworks,Karras2021,garcea2023data,koetzier2024generating,ktena2024generative}. Its potential advantage is to provide additional disease-specific images, but their usefulness must be established through downstream recognition. Generated images may contain artifacts, repeat existing features, or reflect biases in the source data. Even a visually plausible portrait may emphasize features that do little to distinguish its target disorder from other syndromes. More synthetic images therefore need not produce greater recognition gains. The practical question is whether augmentation improves recognition of real patients as more synthetic images are added, and how those gains are distributed across disorders. An increase in overall accuracy would be encouraging, but it would leave open whether sparsely represented disorders also benefit. Evaluating this distribution is central to judging augmentation as a response to limited and uneven training data.

We study disease-specific synthetic oversampling to address this question. The synthetic images come from the Phenotype Diseased Image Synthesis Database (PDIDB)
\citep{pdidb}, which contains facial images generated with our disease-conditioned StyleGAN3 model
\citep{Karras2021}. Its public GitHub repository documents the generation framework and links to a gallery organized by disorder. We progressively add nested portions of these disease-specific pools to a fixed real training set from the GestaltMatcher Database (GMDB)
\citep{lesmann2024gestaltmatcher}. This approach increases the training examples available for underrepresented disorders without requiring additional patient photographs. The nested subsets preserve images as augmentation increases, and fixed patient-disjoint real-data partitions allow conditions to be compared on the same evaluation cohort. Synthetic images are used only for training. We examine GestaltMatcher-Arc and GestaltMML to assess augmentation in image-based and multimodal training settings
\citep{hustinx2023improving,wu2024gestaltmml}. In GestaltMatcher-Arc, oversampling is evaluated alongside class-frequency loss weighting, assessing augmentation within a setup that already addresses imbalance during optimization. The evaluation follows three linked questions: whether augmentation improves aggregate recognition, whether those improvements extend across individual disorders, and whether they coincide with changes in learned representations. We report overall and mean per-class accuracy, followed by disease-specific changes, to distinguish cohort-level gains from their distribution across disorders. An exploratory analysis of class compactness and separation
\citep{rousseeuw1987silhouettes} further examines whether recognition changes coincide with changes in learned representations.

Experiments on a patient-disjoint real-image evaluation set show higher mean recognition accuracy with synthetic augmentation in both training setups. GestaltMatcher-Arc exhibits modest gains and small differences among augmented conditions, whereas GestaltMML continues to improve across the tested levels. These overall gains are uneven across disorders, and the augmentation level with the highest overall accuracy does not always yield the highest mean per-class accuracy. An exploratory GestaltMatcher-Arc embedding analysis also associates fewer real training images with larger gains in class separation, although geometric changes do not consistently track classification performance. Together, these findings support assessing the value of synthetic oversampling through its effects on real-image recognition at both the cohort and disease levels. They also clarify the role of representation analysis as complementary evidence when judging whether additional synthetic images provide useful training information.
\section{Related Work}

\subsection{Deep Facial Phenotyping}

DeepGestalt formulates facial phenotyping as syndrome recognition from photographs
\citep{gurovich2019identifying}. GestaltMatcher learns facial descriptors for similarity-based patient matching in a shared embedding space
\citep{hsieh2022gestaltmatcher}. The GestaltMatcher Database (GMDB) provides a resource for evaluating these representations across genetic disorders
\citep{lesmann2024gestaltmatcher}.

Multimodal methods incorporate information beyond facial appearance. GestaltMML combines facial images with clinical text for rare-disease recognition
\citep{wu2024gestaltmml}, while PhenoGPT2 uses both modalities for phenotype extraction and normalization
\citep{phenogpt2}. Related biomedical work investigates multimodal knowledge transfer through preference optimization
\citep{mint} and rare-disease diagnosis through retrieval-augmented reasoning over clinical notes
\citep{cot-rag}. Our study examines how supplementing facial training data with synthetic images affects image-based and multimodal recognition.

\subsection{Long-Tailed Recognition and Synthetic Oversampling}

Long-tailed recognition methods address unequal class frequencies through different stages of learning and prediction
\citep{liu2019large}. Loss-based approaches include class reweighting
\citep{cui2019class}, class-dependent margins
\citep{cao2019learning}, and focal loss
\citep{Lin_2017_ICCV}. Other approaches decouple representation and classifier training
\citep{kang2019decoupling} or adjust prediction scores through logit adjustment
\citep{menon2021longtaillearninglogitadjustment} and residual correction. REPAIR decomposes this correction into pairwise and classwise components
\citep{repair}. These methods modify the influence of available examples or their prediction scores, but do not provide additional observations of underrepresented classes.

Synthetic oversampling supplements the training data with generated examples
\citep{goodfellow2014generativeadversarialnetworks,Karras2021}. Its use has been investigated in medical imaging
\citep{garcea2023data,koetzier2024generating,ktena2024generative}, and GestaltGAN applies generative modeling to rare-disease facial portraits
\citep{kirchhoff2025gestaltgan}. In this study, PDIDB
\citep{pdidb} supplies synthetic faces generated with our disease-conditioned StyleGAN3 model
\citep{Karras2021}. Synthetic artifacts, repeated features, and source-data biases can limit the usefulness of these additional examples
\citep{garcea2023data,koetzier2024generating}. We assess this usefulness by comparing nested augmentation levels on a fixed real-image evaluation partition, using disease-specific accuracy and embedding geometry to complement aggregate performance.
\section{Methods}
\label{sec:methods}
\subsection{Study Design and Dataset}

\begin{figure}[htbp]
    \centering
    \includegraphics[width=0.90\textwidth]
    {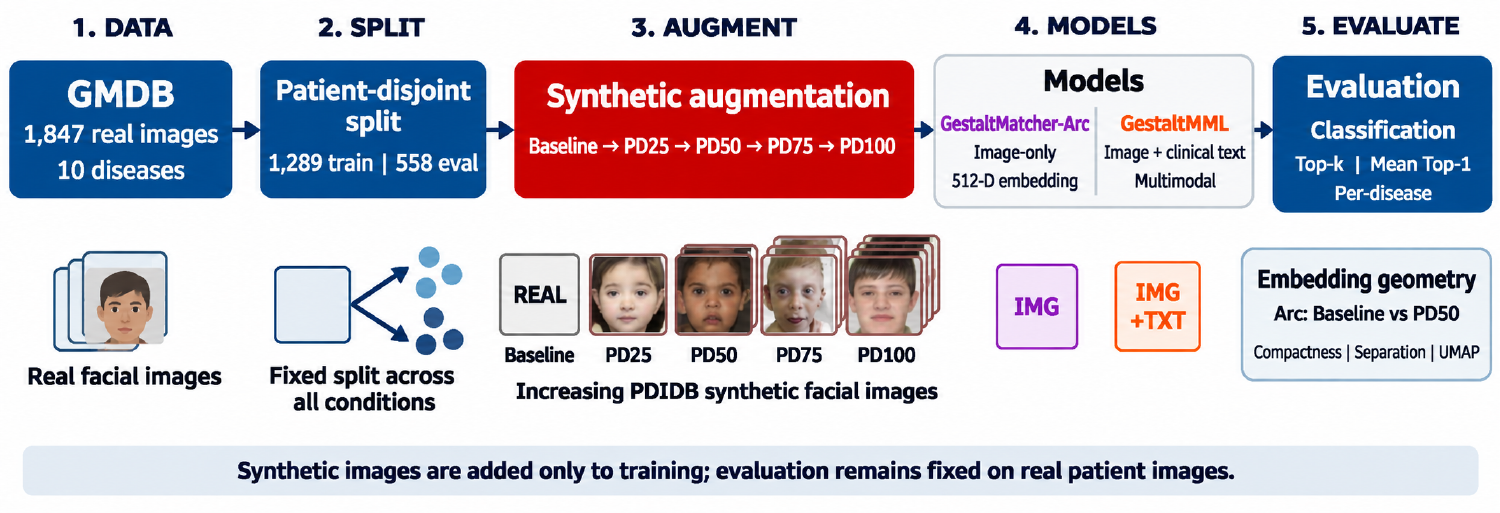}
    \caption{Study design. A patient-disjoint GMDB partition contains
    1,289 real training images and 558 real evaluation images. Nested
    PDIDB subsets augment the training data under PD25, PD50, PD75,
    and PD100. GestaltMatcher-Arc and GestaltMML are trained with three
    seeds per condition. The real partitions and synthetic subsets are
    fixed across training seeds.}
    \label{fig:workflow}
\end{figure}

We used a curated subset of the GestaltMatcher Database (GMDB v1.1.0)
\citep{lesmann2024gestaltmatcher} containing 1,847 real facial images from
10 genetic disorders. A fixed patient-disjoint partition assigned 1,289
images to training and 558 images to evaluation, corresponding to an
approximately 70\%/30\% split. Training counts ranged from 312 images for
Cornelia de Lange syndrome to 41 images for 22q11.2 deletion syndrome,
giving a largest-to-smallest class ratio of 7.6. The real-data partitions
were identical across models, augmentation conditions, and training seeds.
Synthetic images were added only to training. Figure~\ref{fig:workflow}
summarizes the study design.

To screen for potential overlap between synthetic training images and the real evaluation partition, we compared the full synthetic pool against all evaluation images using exact pixel-level hashes, perceptual hashes, and nearest-neighbor cosine similarity in the 512-dimensional GestaltMatcher-Arc embedding space. Embedding similarities were calibrated against real-training-to-evaluation nearest-neighbor similarities.

\subsection{Synthetic Image Augmentation}

We obtained synthetic facial images from the Phenotype Diseased Image
Synthesis Database (PDIDB) \citep{pdidb}, generated with our
disease-conditioned StyleGAN3 model \citep{Karras2021}. We defined five experimental conditions. 
Baseline used only
real training images. PD25, PD50, PD75, and PD100 added approximately
25\%, 50\%, 75\%, and 100\% of each disorder's available synthetic pool,
respectively. These percentages refer to the synthetic pool, not to the
size of the real training set.

\begin{figure}[htbp]
    \centering
    \includegraphics[width=0.82\textwidth]
    {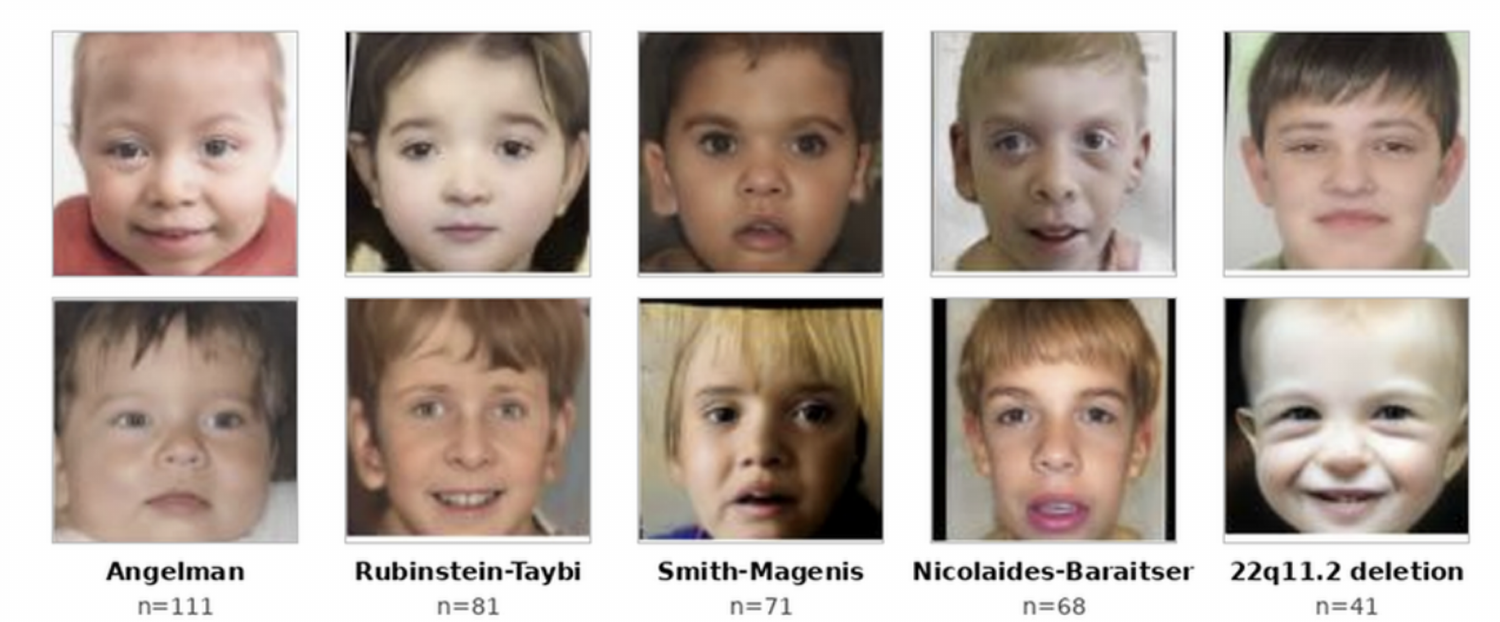}
    \caption{PDIDB synthetic facial images. Two examples are shown for
    each of the five disorders with the fewest real training images.
    Synthetic images are used only for training.}
    \label{fig:synthetic_examples}
\end{figure}

The synthetic subsets were sampled once using random seed 11 and nested across augmentation conditions. These subsets were held fixed across all model-training runs, so the three training seeds did not change which synthetic images were included. Each higher augmentation condition therefore retained the images in the preceding condition and added further samples. This design compares increasing amounts from one fixed synthetic pool while avoiding variation caused by independently resampling each condition.
Table~\ref{tab:dataset_composition} reports
the real and synthetic image counts. The augmented conditions added 433,
866, 1,298, and 1,727 images, producing total training sets of 1,722,
2,155, 2,587, and 3,016 images. Figure~\ref{fig:synthetic_examples} shows
examples from the disorders with the fewest real training images.

\begin{table}[htbp]
    \centering
    \caption{Dataset composition. Train and Eval. are the fixed,
    patient-disjoint real-image partitions. The Baseline condition consists of the real training images only,  while PD25--PD100 indicate the numbers of synthetic images added under the four augmentation conditions.}
    \label{tab:dataset_composition}
    \small
    \setlength{\tabcolsep}{2pt}
    \begin{tabular}{lrrrrrr}
        \toprule
        \textbf{Disorder} &
        \textbf{Train} &
        \textbf{Eval.} &
        \textbf{PD25} &
        \textbf{PD50} &
        \textbf{PD75} &
        \textbf{PD100} \\
        \midrule
        Cornelia de Lange     & 312 & 135 & 46 & 91 & 137 & 182 \\
        Williams-Beuren       & 191 & 82  & 47 & 93 & 140 & 186 \\
        Noonan                & 145 & 62  & 48 & 95 & 143 & 190 \\
        Kabuki                & 144 & 62  & 40 & 81 & 121 & 161 \\
        KBG                   & 125 & 54  & 47 & 94 & 140 & 187 \\
        Angelman              & 111 & 49  & 42 & 85 & 127 & 169 \\
        Rubinstein-Taybi      & 81  & 35  & 44 & 89 & 133 & 177 \\
        Smith-Magenis         & 71  & 31  & 42 & 83 & 125 & 166 \\
        Nicolaides-Baraitser  & 68  & 29  & 42 & 84 & 126 & 168 \\
        22q11.2 deletion      & 41  & 19  & 35 & 71 & 106 & 141 \\
        \midrule
        \textbf{Total} &
        \textbf{1,289} &
        \textbf{558} &
        \textbf{433} &
        \textbf{866} &
        \textbf{1,298} &
        \textbf{1,727} \\
        \bottomrule
    \end{tabular}
\end{table}

\subsection{Model Training}
\label{sec:model_training}
Both models were trained under all five conditions using three model-training random seeds (11, 22, and 33) on NVIDIA A100 GPUs. These seeds controlled training stochasticity, including random parameter initialization and data shuffling, while the real-data partitions and synthetic subsets remained unchanged.

\subsubsection{GestaltMatcher-Arc}

GestaltMatcher-Arc \citep{hustinx2023improving} uses a ResNet-50 face
backbone pretrained on Glint360K \citep{an2021partial} with ArcFace loss
\citep{deng2019arcface}. We froze the convolutional backbone and trained
the 512-dimensional feature projection and 10-class classifier for
50 epochs. Training used Adam optimizer with a batch size of 64 and learning rates
of $5\times10^{-4}$ for the projection layer and $10^{-3}$ for the classifier.
A ReduceLROnPlateau scheduler used a reduction factor of 0.5 and patience
of five epochs.

We used class-frequency-weighted cross-entropy with class weight
\begin{equation}
    w_c = 0.5 + 0.5\frac{n_{\min}}{n_c},
    \label{eq:class_weight}
\end{equation}
where $n_c$ denotes the training count for class $c$ and $n_{\min}$ is
the smallest class count. Synthetic oversampling was thus evaluated
together with loss weighting. To assess whether the augmentation effect depended on class-frequency weighting, we repeated the Baseline and PD50 conditions across all three seeds using unweighted cross-entropy as an ablation study.

\subsubsection{GestaltMML}

We initialized GestaltMML \citep{wu2024gestaltmml} from
\texttt{dandelin/vilt-b32-mlm} \citep{kim2021vilt} and used a 10-class
output head. Each real training image formed two image-text pairs.
One contained clinical text with sex, age, ethnicity, and Human Phenotype
Ontology (HPO) terms; the other used the placeholder token ``*''.
Each synthetic image formed one pair with the placeholder because
corresponding clinical metadata were unavailable. This construction
produced 2,578, 3,011, 3,444, 3,876, and 4,305 training pairs for
Baseline, PD25, PD50, PD75, and PD100, respectively.

We fine-tuned the model for 10 epochs using AdamW with a
learning rate of $5\times10^{-5}$, a batch size of 64, and a maximum
text length of 40.
During evaluation, each real image was paired once with its own clinical text, constructed from the same demographic fields and HPO terms used for the matched real training pairs. Missing demographic fields were represented as ``unknown''. The placeholder-only copies were used for training and were not additional evaluation examples.

\subsection{Classification Evaluation}
\label{sec:metric_definition}

For each training run, we selected the checkpoint with the highest overall Top-1 accuracy on the real-image evaluation partition, breaking ties in favor of the earliest epoch. The same partition was used to report classification performance.

Let $N$ be the number of evaluation images, $C=10$ the number of classes,
and $N_c$ the number of evaluation images in class $c$. For image $i$,
$y_i$ is the true label and $\hat{y}_i$ is the predicted label. Overall
Top-1 accuracy weights images equally:
\begin{equation}
    \mathrm{Top1}
    = \frac{1}{N}\sum_{i=1}^{N}\mathbf{1}(\hat{y}_i=y_i).
    \label{eq:top1}
\end{equation}
Mean per-class Top-1 accuracy weights classes equally:
\begin{equation}
    \mathrm{MeanTop1}
    = \frac{1}{C}\sum_{c=1}^{C}
      \frac{1}{N_c}\sum_{i:y_i=c}\mathbf{1}(\hat{y}_i=c).
    \label{eq:mean_top1}
\end{equation}
Reporting both Top-1 summaries distinguishes performance on the observed image distribution from performance averaged equally across disorders. In this closed-set task, predictions are restricted to the ten study disorders.
We also report Top-3 and Top-5 accuracy, defined by whether the true
label appears among the three or five highest-ranked predictions.

Classification metrics are reported as mean and sample standard
deviation across the three training seeds. For disease-specific
comparisons, we calculated the percentage-point difference from
Baseline within each seed and then summarized these paired differences
across seeds.

\subsection{Embedding Analysis}

We conducted an exploratory analysis of GestaltMatcher-Arc embeddings
using the seed-11 Baseline and PD50 runs. PD50 was selected after
examining classification results because it had the highest mean
overall Top-1 accuracy across training seeds. Both runs were analyzed
on the same real evaluation images using their $L_2$-normalized,
512-dimensional embeddings.

Global class structure was summarized by the mean cosine-distance
silhouette score \citep{rousseeuw1987silhouettes}. For image $i$,
\begin{equation}
    s(i) = \frac{b(i)-a(i)}{\max\{a(i),b(i)\}},
    \label{eq:silhouette}
\end{equation}
where $a(i)$ is the mean cosine distance to other images in the same
class. For each other class, we calculated the mean cosine distance
from image $i$ to that class; $b(i)$ is the smallest of these means.

We additionally calculated three disease-level measures. Compactness
is the mean cosine distance between a class's embeddings and its
$L_2$-normalized centroid. Nearest-centroid separation is the cosine
distance from that centroid to the closest centroid of another class.
The separation-to-compactness ratio divides separation by compactness.
Lower compactness indicates tighter clusters, while higher separation
indicates greater distance between class centroids. All quantitative
measures were calculated in the original embedding space. PCA and UMAP
were used only for visualization.

\subsection{Statistical Analysis}

For each embedding metric $M$, we calculated the relative change from
Baseline to PD50 as
\begin{equation}
    \Delta M(\%)
    = 100\frac{M_{\mathrm{PD50}}-M_{\mathrm{Baseline}}}
    {M_{\mathrm{Baseline}}}.
    \label{eq:relative_change}
\end{equation}
We used Spearman rank correlation with two-sided $p$-values to assess
associations between real training class size and relative changes in
each metric. These analyses involved 10 disorders and a single training
seed and were interpreted as exploratory associations.
\FloatBarrier
\section{Results}
\label{sec:experiments}

We evaluate synthetic augmentation on the same real-image cohort at each level. After screening for detectable overlap between synthetic and evaluation images, we assess whether aggregate recognition improves, how the gains are distributed across disorders, and whether augmentation is accompanied by changes in representation geometry.

\subsection{Experimental Setup}

We compare GestaltMatcher-Arc and GestaltMML across Baseline and four nested synthetic augmentation conditions (PD25--PD100), using the fixed patient-disjoint partition described in Section~\ref{sec:methods}. Each condition was trained with seeds 11, 22, and 33. We compare augmentation effects within each model because their architectures and training procedures differ.

\subsection{Synthetic--Evaluation Overlap Screening}

Screening all 1,727 synthetic images against the 558 real evaluation images identified no exact pixel duplicates. Median nearest-neighbor cosine similarity was 0.591 for synthetic-to-evaluation comparisons and 0.670 for the patient-disjoint real-training-to-evaluation reference; the corresponding maxima were 0.851 and 0.898. Manual inspection of the closest perceptual-hash and embedding matches found no apparent duplicate images. These checks address detectable image duplication rather than establishing independence from the generator's original training cohort. Aggregate audit summaries are available in the code repository.

\subsection{Overall Classification Performance}

\begin{figure}[htbp]
    \centering
    \includegraphics[width=0.92\textwidth]
    {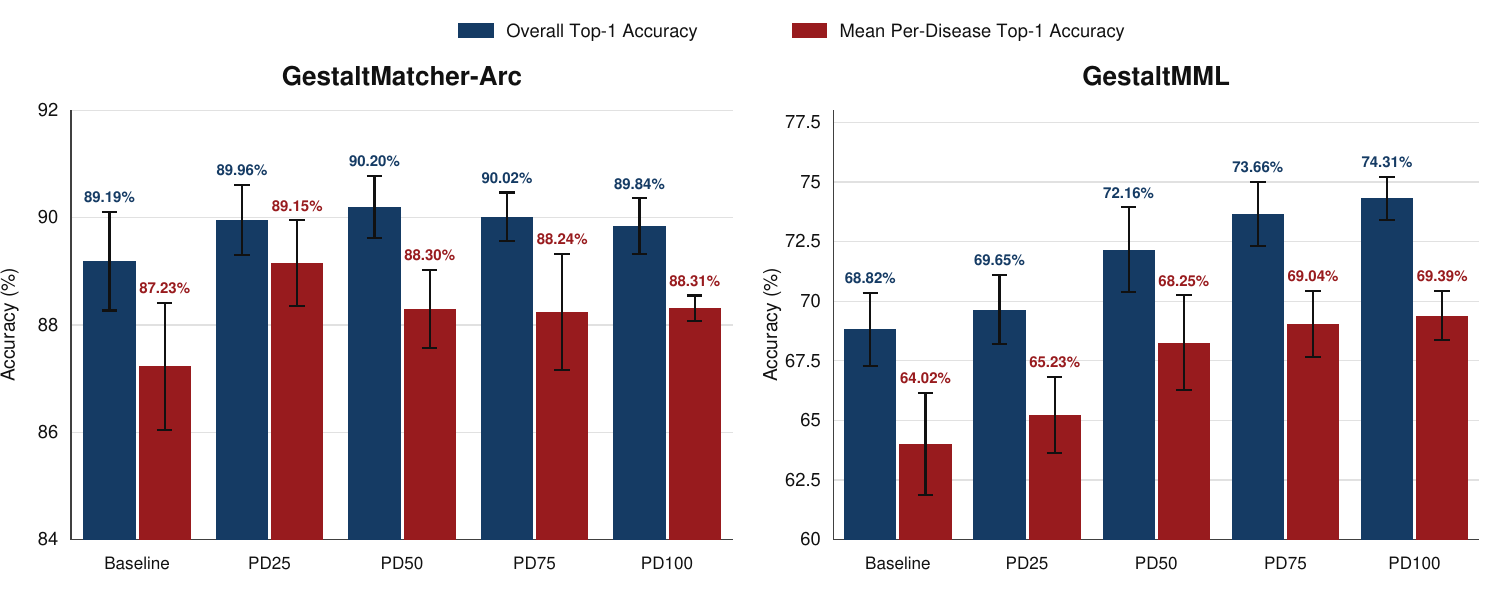}
    \caption{Classification accuracy across augmentation conditions.
    Bars show mean overall and mean per-class Top-1 accuracy across
    three training seeds; error bars show one sample standard deviation.
    Checkpoints are selected and evaluated on the same fixed real-image
    partition.}
    \label{fig:overall_performance}
\end{figure}

Figure~\ref{fig:overall_performance} compares overall and mean per-class
Top-1 accuracy across augmentation conditions. Both models achieved
higher mean overall Top-1 accuracy with synthetic augmentation than at
Baseline, but their responses to additional synthetic images differed.

For GestaltMatcher-Arc, mean overall Top-1 accuracy increased from
$89.19\pm0.92\%$ at Baseline to its highest observed value of
$90.20\pm0.58\%$ at PD50. Differences among augmented conditions were
small relative to the variation across training seeds. Mean per-class
Top-1 accuracy instead peaked at PD25, reaching $89.15\pm0.80\%$ compared
with $87.23\pm1.18\%$ at Baseline. Thus, the condition with the highest
overall accuracy differed from the condition with the highest mean
per-class accuracy. 

To assess sensitivity to class-frequency weighting, we repeated the Arc Baseline and PD50 conditions using unweighted cross-entropy (Table~S2). The paired PD50 gain was $+1.02 \pm 1.72$ percentage points without weighting, compared with $+1.02 \pm 1.49$ points with weighting. These almost identical mean gains suggest that the observed augmentation effect did not depend strongly on class-frequency weighting.

For GestaltMML, mean overall Top-1 accuracy increased across the tested
augmentation levels, from $68.82\pm1.53\%$ at Baseline to
$74.31\pm0.90\%$ at PD100. Mean per-class Top-1 accuracy also increased,
from $64.02\pm2.14\%$ to $69.39\pm1.03\%$. Complete Top-1, mean per-class
Top-1, Top-3, and Top-5 results are reported in
Table~S1 in the Supporting Information.


\subsection{Aggregate Gains Include Uneven Disease-Level Responses}

\begin{figure}[p]
    \centering
    \newcommand{\HeatmapCrop}[1]{%
        \includegraphics[height=0.381\textheight,trim=#1,clip]%
        {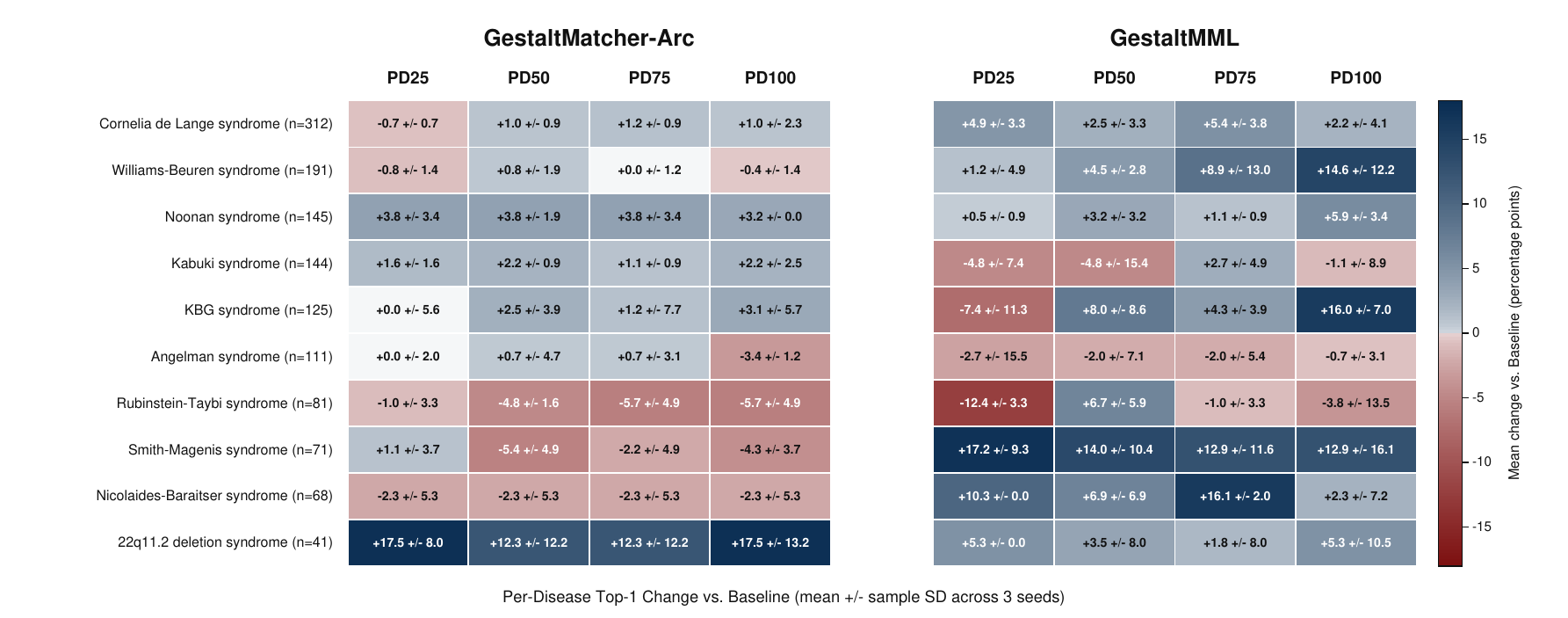}%
    }
    \begin{minipage}[c]{0.86\linewidth}
        \centering\singlespacing
        \makebox[\linewidth][l]{\sffamily\bfseries A\quad GestaltMatcher-Arc}\par
        \vspace{1mm}
        \HeatmapCrop{45.69bp 23.80bp 399.77bp 29.51bp}\par
        \vspace{5mm}
        \makebox[\linewidth][l]{\sffamily\bfseries B\quad GestaltMML}\par
        \vspace{1mm}
        \HeatmapCrop{45.69bp 23.80bp 667.24bp 29.51bp}%
        \HeatmapCrop{509.24bp 23.80bp 80.91bp 29.51bp}
    \end{minipage}%
    \hspace{2mm}%
    \raisebox{-0.5\height}{\includegraphics[height=0.381\textheight,
        trim=780.51bp 23.80bp 20.94bp 29.51bp,clip]%
        {liao_manuscript/figures/figure4_disease_specific_heatmaps_multiseed.pdf}}
    \caption{Disease-specific changes in Top-1 accuracy from Baseline.
    Panel A shows GestaltMatcher-Arc and panel B shows GestaltMML.
    Values are mean $\pm$ sample standard deviation of paired
    percentage-point differences across three training seeds. Each
    difference is calculated against the Baseline run with the same seed.
    Disorders are ordered by real training image count, from largest
    to smallest.}
    \label{fig:disease_heatmaps}
\end{figure}

The different Arc peaks for overall and mean per-class accuracy motivate a closer examination of individual disorders. We next ask whether the aggregate gains reflect broadly shared improvements or a mixture of gains and losses. Figure~\ref{fig:disease_heatmaps} reports paired changes in disease-specific
Top-1 accuracy from Baseline, calculated within each training seed and
then averaged across seeds. For GestaltMatcher-Arc, higher mean overall
accuracy did not imply improvement for every disorder; disease-specific
changes varied across augmentation conditions and training seeds.

At PD100, GestaltMML improved mean Top-1 accuracy for seven disorders.
The largest paired gains occurred for KBG ($+16.05$ percentage points),
Williams-Beuren ($+14.63$), and Smith-Magenis ($+12.90$). Mean accuracy
also increased for Noonan ($+5.91$), 22q11.2 deletion ($+5.26$),
Nicolaides-Baraitser ($+2.30$), and Cornelia de Lange ($+2.22$).
In contrast, mean accuracy decreased for Rubinstein-Taybi
($-3.81$ percentage points), Kabuki ($-1.08$), and Angelman ($-0.68$).
The improvement in aggregate performance therefore included both gains
and losses at the disease level.

\subsection{Embedding-Space Analysis}

To examine whether augmentation also changes the organization of disease representations, we compared Baseline and PD50 embeddings from the seed-11
GestaltMatcher-Arc runs using the same 558 real evaluation images.
PD50 was selected because it achieved the highest mean overall Top-1
accuracy across seeds; this analysis is exploratory. The mean
cosine-distance silhouette score increased from 0.228 at Baseline to
0.267 at PD50. This corresponds to a relative increase of 16.9\%,
calculated from unrounded values.

Nearest-centroid separation increased for nine of the ten disorders
(Table~S3). The largest relative increases were
observed for Smith-Magenis and 22q11.2 deletion (both $+23.91\%$),
followed by KBG ($+18.56\%$) and Angelman ($+18.46\%$). Cornelia de Lange
was the only disorder with reduced separation ($-9.86\%$).

Compactness improved for eight disorders. Mean distance to the class
centroid increased slightly for Smith-Magenis ($+2.35\%$) and 22q11.2
deletion ($+0.94\%$), even though their nearest-centroid separation
increased. The separation-to-compactness ratio increased for nine
disorders and decreased for Cornelia de Lange ($-0.56\%$). These results
show that compactness and separation can change differently under the
same augmentation condition.

\begin{figure}[p]
    \centering
    \newcommand{\UMAPArtwork}{liao_manuscript/figures/figure5a_arc_umap_baseline_vs_pd50.pdf}
    \begin{minipage}[c]{0.7075\linewidth}
        \centering\singlespacing
        \makebox[\linewidth][l]{\sffamily\bfseries A\quad Baseline}\par
        \vspace{1mm}
        \includegraphics[height=0.3585\textheight,trim=0bp 0bp 422.4bp 22.8bp,clip]{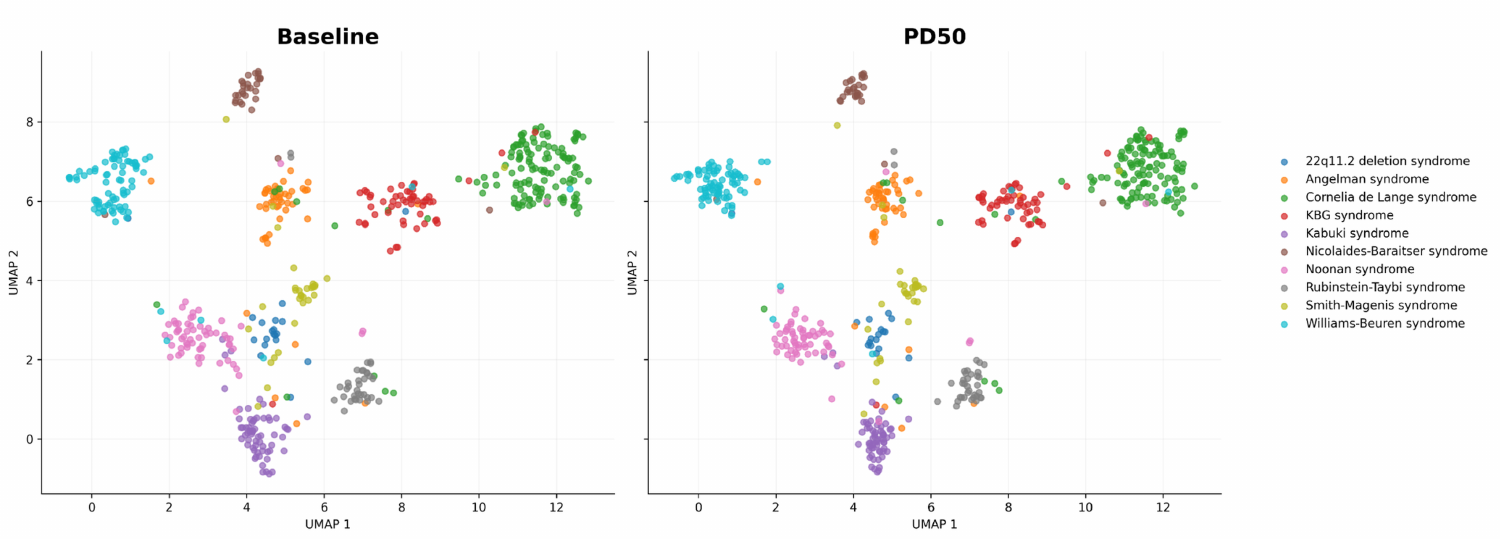}
        \par\vspace{5mm}
        \makebox[\linewidth][l]{\sffamily\bfseries B\quad PD50}\par
        \vspace{1mm}
        \hspace*{2.7mm}%
        \includegraphics[height=0.3585\textheight,trim=297.6bp 0bp 132bp 22.8bp,clip]{\UMAPArtwork}
    \end{minipage}%
    \hfill
    \begin{minipage}[c]{0.28\linewidth}
        \raggedright\sffamily\singlespacing
        \newcommand{\DisorderKey}[2]{%
            \makebox[1em][l]{\textcolor[RGB]{#1}{\textbullet}}%
            #2\par\vspace{0.7mm}%
        }
        {\bfseries Disorder}\par\vspace{2mm}
        \fontsize{9}{11}\selectfont
        \setlength{\parskip}{0pt}
        \DisorderKey{104,170,209}{22q11.2 deletion syndrome}
        \DisorderKey{255,173,87}{Angelman syndrome}
        \DisorderKey{104,190,107}{Cornelia de Lange syndrome}
        \DisorderKey{230,105,100}{KBG syndrome}
        \DisorderKey{190,163,215}{Kabuki syndrome}
        \DisorderKey{183,144,136}{Nicolaides-Baraitser syndrome}
        \DisorderKey{238,169,216}{Noonan syndrome}
        \DisorderKey{168,168,168}{Rubinstein-Taybi syndrome}
        \DisorderKey{214,212,111}{Smith-Magenis syndrome}
        \DisorderKey{95,213,226}{Williams-Beuren syndrome}
    \end{minipage}
    \caption{Shared UMAP projection of seed-11 GestaltMatcher-Arc
    embeddings at Baseline (A) and PD50 (B). Each point represents one of the
    same 558 real evaluation images; colors indicate disorder.
    Quantitative metrics use the original $L_2$-normalized,
    512-dimensional embeddings.}
    \label{fig:arc_umap}
\end{figure}

Figure~\ref{fig:arc_umap} shows a shared UMAP projection of the Baseline
and PD50 representations. Quantitative results are calculated in the
original embedding space.

\subsection{Association With Real Training Class Size}

We next examined whether the geometric changes were related to the amount of real training data available for each disorder. Across the ten disorders, real training class size was inversely
associated with relative change in nearest-centroid separation
(Spearman $\rho=-0.707$, $p=0.022$). Disorders with fewer real training
images tended to show larger relative separation gains at PD50 in
this seed-11 analysis.

The corresponding correlations were $\rho=0.527$ ($p=0.117$) for
relative compactness change and $\rho=-0.588$ ($p=0.074$) for relative
change in the separation-to-compactness ratio. These exploratory
associations describe variation across disorders in one training run;
they do not establish that smaller class size causes a greater benefit
from augmentation.

\FloatBarrier
\section{Discussion}

Synthetic oversampling produced higher mean recognition accuracy in both evaluated training setups, but its value depended on the outcome examined. Aggregate gains coexisted with disease-specific losses, and the Arc augmentation level with the highest overall accuracy differed from that with the highest mean per-class accuracy. Representation analysis added a further distinction: increased class separation did not consistently track recognition performance. Together, these findings show why evaluating synthetic data for rare-disease facial phenotyping requires attention to both the magnitude and the distribution of recognition gains.

The two training setups differed in their response to augmentation amount. GestaltMatcher-Arc showed modest gains with little variation among augmented conditions, while GestaltMML continued to improve across the tested levels. This suggests assessing augmentation amount within each training setup. For Arc, adding the full available pool did not yield the highest mean accuracy; for GestaltMML, PD100 was the largest pool evaluated, so the increasing trend does not establish an optimal amount beyond that range. The similar mean Arc gains with and without class-frequency weighting further suggest that the observed augmentation effect was not strongly dependent on this weighting choice in the tested comparison. These results do not, however, isolate the contribution of additional image variation from other changes accompanying augmentation.

Disease-level outcomes are particularly relevant to the motivation for synthetic oversampling. Overall accuracy reflects the disease frequencies in the evaluation cohort, whereas mean per-class accuracy gives a small cohort the same weight as a large one. The different Arc peaks at PD50 and PD25 therefore reflect different evaluation objectives. Neither aggregate metric guarantees improvement for every disorder: even at the GestaltMML condition with the highest aggregate accuracy, seven disorders improved and three declined. For a disorder whose accuracy declines, an improvement in the cohort average does not represent a recognition benefit. Reporting these paired disease-level changes makes that distinction visible and helps identify where augmentation needs further scrutiny. The present results do not determine whether individual losses arise from synthetic image characteristics, changes in optimization, or variability associated with limited disease-specific samples.

The representation analysis provides a complementary view of this uneven response. The association between smaller real training classes and larger gains in centroid separation is consistent with the possibility that data availability matters for geometric changes. It does not establish greater recognition benefit for those disorders. Cornelia de Lange improves in mean classification accuracy under some augmentation conditions while its separation decreases in the seed-11 PD50 analysis. Moreover, compactness and separation can move in different directions within the same disorder. Centroid-based metrics summarize average class structure, whereas recognition depends on the placement of individual images relative to the learned classifier. A larger centroid distance can coexist with errors among overlapping or atypical examples. Geometry can therefore help characterize what changes after augmentation, but it cannot substitute for disease-specific recognition outcomes or explain accuracy changes across repeated training runs on its own.

Several design features limit the interpretation of these findings. The models differ in backbones, trainable parameters, optimization, and input construction, so their different responses cannot be attributed to a particular component or to multimodality alone. Fixed epoch counts give larger augmented datasets more optimizer updates. In GestaltMML, real images were paired with both clinical and placeholder text, whereas synthetic images were paired only with placeholder text; adding synthetic pairs therefore also changes the proportion of clinical-text pairs. These factors remain coupled to augmentation, limiting attribution of the gains solely to synthetic image quantity.

The evaluation also covers ten disorders, one synthetic image source, and one patient partition. Repeated training seeds capture training variability but do not assess uncertainty from alternative patient partitions or synthetic subsets. The representation analysis uses one seed and a condition selected after examining classification results, limiting the conclusions from its cross-disease associations. Checkpoint selection and performance reporting use the same evaluation partition, so the reported accuracies are not independent test estimates. The overlap screening addresses detectable duplication without establishing independence from the generator's original training cohort. The findings therefore concern recognition on this benchmark; clinical utility has not been established.

\section{Conclusion}

Disease-specific synthetic oversampling improved mean recognition accuracy on this ten-disorder benchmark, with modest gains for GestaltMatcher-Arc and continued gains across the tested augmentation levels for GestaltMML. The central finding is that these aggregate improvements were uneven across disorders, and the augmentation level favored by overall accuracy could differ from that favored by mean per-class accuracy. Evaluating synthetic data therefore requires examining which disorders benefit alongside the cohort average. Exploratory representation analysis adds information about class structure, but does not establish the mechanism underlying recognition changes.

Further evaluation should use separate validation and test partitions, external patient cohorts, and alternative synthetic subsets and generators to assess the stability of both aggregate and disease-specific findings. Comparisons with matched training budgets and image--text ablations would help distinguish the effects of additional images from changes in optimization and input composition. Cross-modal attribution methods such as MultiSHAP \citep{multishap} could complement these comparisons by characterizing the contributions of facial images, clinical text, and their interactions to model predictions. Such evaluations would clarify when additional synthetic examples provide a reproducible benefit for recognition of real patients.

\FloatBarrier
\section*{Supporting Information}
\ifdefined\IncludeSupportingInformation
Supporting Information follows the References and contains complete classification results
(Table~\ref{tab:overall_results}), the class-weighting sensitivity analysis
(Table~\ref{tab:class_weight_ablation}), and disease-specific embedding metrics
(Table~\ref{tab:embedding_results}).
\else
Supporting Information is provided as a separate document and contains complete classification results (Table S1), the class-weighting sensitivity analysis (Table S2), and disease-specific embedding metrics (Table S3).
\fi
\section*{Acknowledgements}
We thank the members of the Wang Genomics Lab at the University of
Pennsylvania and Children's Hospital of Philadelphia for hosting this research and providing guidance, feedback, and technical support. We also thank Dr. Priscilla Encarnacao, Mrs. Kristen Abraham, and Dr. Early Geyer III for supporting the summer research experience through the Hopkins Authentic Research Program in Science (HARPS). We acknowledge the creators and contributors of the
GestaltMatcher Database (GMDB) and the PDIDB synthetic dataset for
making these resources available.
Synthetic image generation was supported by the National Institutes of Health under grant
OD037960 and by the CHOP Research Institute.
\section*{Funding}
This study is supported in part by the Children's Hospital of Philadelphia.
\section*{Conflict of Interest}
The authors declare no competing interests.
\section*{Ethics Approval and Consent}
This study used existing GMDB data and PDIDB synthetic images; no new patients were recruited. GMDB reports broad participant consent and approval from the Ethics Committee of the Medical Faculty of the University of Bonn (reference 386/17). Patient or guardian consent for data sharing is managed through GMDB. The facial portraits presented in Figure~\ref{fig:synthetic_examples} are synthetic.
\section*{Data and Code Availability}
Experimental details, including data partitioning, augmentation conditions,
model training, evaluation, and statistical analyses, are provided in the
Methods and Supporting Information. Code for training, evaluation, and analysis is
available in our
\href{https://github.com/mliao2710/synthetic-oversampling-facial-phenotyping}
{GitHub repository}.
The repository also provides experiment configurations and aggregate
results; restricted patient data and model checkpoints are not redistributed.
GMDB v1.1.0 is available to researchers on application from the GestaltMatcher Database (\url{https://db.gestaltmatcher.org/}), subject to approval by its Scientific Advisory Board and the applicable data-use conditions. Real patient images and restricted clinical data are not redistributed with this article. The PDIDB generation framework and a link to its synthetic-image gallery are available at \url{https://github.com/WGLab/PDIDB}; use is subject to the resource's stated terms. Exact public PDIDB image identifiers for the nested augmentation conditions are provided in the analysis repository under \texttt{data/pdidb\_manifests/}.
\section*{AI Assistance Disclosure}
OpenAI ChatGPT 5.6 Sol and Anthropic Claude Sonnet 5.5 were used during manuscript preparation to assist with language editing concentrated in the Appendix, figures, and figure legends. The same models were also used for organization, LaTeX formatting, and code development and debugging. All generated material was reviewed and revised by the authors. The authors independently verified the experimental results, statistical analyses, references, and scientific claims and take responsibility for the final manuscript. ChatGPT was additionally used to assist with journal-format adaptation and preparation of the revised abstract and cover-letter draft, and clarification of the methods, results, and discussion. These revisions require author review before submission.
\printbibliography[title={References}]
\ifdefined\IncludeSupportingInformation
  \clearpage
  {\Large\bfseries Supporting Information\par}
  \setcounter{section}{0}
\setcounter{table}{0}
\setcounter{figure}{0}
\renewcommand{\thesection}{S\arabic{section}}
\renewcommand{\thetable}{S\arabic{table}}
\renewcommand{\thefigure}{S\arabic{figure}}
\renewcommand{\theHsection}{supp.\arabic{section}}
\renewcommand{\theHtable}{supp.\arabic{table}}
\renewcommand{\theHfigure}{supp.\arabic{figure}}
\captionsetup{hypcap=false}

\section{Complete Classification Results}
\begin{table}[h]
\centering
\caption{
Overall classification performance across synthetic augmentation conditions.
Top-$k$ denotes sample-weighted Top-$k$ accuracy, while Mean Top-1 denotes
Top-1 accuracy averaged equally across the 10 disease classes. Values are
mean $\pm$ sample SD across three training seeds, in percentages. Green
indicates the highest-observed mean and red indicates the lowest-observed
mean within each architecture and metric.
}
\label{tab:overall_results}

\small
\setlength{\tabcolsep}{3pt}

\resizebox{0.96\columnwidth}{!}{%
\begin{tabular}{llrrrr}
\toprule
\textbf{Model} &
\textbf{Condition} &
\textbf{Top-1} &
\textbf{Mean Top-1} &
\textbf{Top-3} &
\textbf{Top-5} \\
\midrule

\multirow{5}{*}{GestaltMatcher-Arc}
& Baseline
    & \cellcolor{worstresult}89.19 $\pm$ 0.92
    & \cellcolor{worstresult}87.23 $\pm$ 1.18
    & 97.25 $\pm$ 0.27
    & \cellcolor{worstresult}98.27 $\pm$ 0.27 \\
& PD25
    & 89.96 $\pm$ 0.65
    & \cellcolor{bestresult}\textbf{89.15 $\pm$ 0.80}
    & \cellcolor{worstresult}97.01 $\pm$ 0.27
    & 98.45 $\pm$ 0.37 \\
& PD50
    & \cellcolor{bestresult}\textbf{90.20 $\pm$ 0.58}
    & 88.30 $\pm$ 0.73
    & \cellcolor{bestresult}\textbf{97.61 $\pm$ 0.10}
    & \cellcolor{bestresult}\textbf{98.86 $\pm$ 0.21} \\
& PD75
    & 90.02 $\pm$ 0.45
    & 88.24 $\pm$ 1.08
    & 97.25 $\pm$ 0.37
    & 98.69 $\pm$ 0.27 \\
& PD100
    & 89.84 $\pm$ 0.52
    & 88.31 $\pm$ 0.24
    & 97.13 $\pm$ 0.36
    & 98.63 $\pm$ 0.10 \\
\midrule

\multirow{5}{*}{GestaltMML}
& Baseline
    & \cellcolor{worstresult}68.82 $\pm$ 1.53
    & \cellcolor{worstresult}64.02 $\pm$ 2.14
    & \cellcolor{worstresult}86.26 $\pm$ 0.83
    & \cellcolor{worstresult}93.55 $\pm$ 1.12 \\
& PD25
    & 69.65 $\pm$ 1.45
    & 65.23 $\pm$ 1.58
    & 87.28 $\pm$ 0.47
    & 94.03 $\pm$ 0.45 \\
& PD50
    & 72.16 $\pm$ 1.77
    & 68.25 $\pm$ 1.99
    & 88.41 $\pm$ 1.63
    & 94.15 $\pm$ 0.99 \\
& PD75
    & 73.66 $\pm$ 1.35
    & 69.04 $\pm$ 1.39
    & 89.55 $\pm$ 1.08
    & \cellcolor{bestresult}\textbf{95.40 $\pm$ 0.58} \\
& PD100
    & \cellcolor{bestresult}\textbf{74.31 $\pm$ 0.90}
    & \cellcolor{bestresult}\textbf{69.39 $\pm$ 1.03}
    & \cellcolor{bestresult}\textbf{89.84 $\pm$ 0.55}
    & 95.34 $\pm$ 1.00 \\
\bottomrule
\end{tabular}%
}

\end{table}
\FloatBarrier
\section{Class-Weighting Sensitivity Analysis}
\begin{center}

\captionof{table}{
GestaltMatcher-Arc sensitivity to class-frequency weighting. Values are mean
$\pm$ sample SD across three training seeds. $\Delta$ denotes the paired
seed-wise PD50 minus Baseline difference in overall Top-1 accuracy.
}
\label{tab:class_weight_ablation}

\vspace{2pt}

\scriptsize
\setlength{\tabcolsep}{3pt}
\renewcommand{\arraystretch}{1.05}

\resizebox{\columnwidth}{!}{%
\begin{tabular}{lccc}
\toprule
\textbf{Loss} &
\textbf{Baseline Top-1 (\%)} &
\textbf{PD50 Top-1 (\%)} &
\textbf{$\Delta$ (pp)} \\
\midrule

Weighted CE
& $89.19 \pm 0.92$
& $90.20 \pm 0.58$
& $+1.02 \pm 1.49$ \\

Unweighted CE
& $89.25 \pm 1.00$
& $90.26 \pm 0.72$
& $+1.02 \pm 1.72$ \\

\bottomrule
\end{tabular}%
}

\end{center}
\FloatBarrier
\clearpage
\section{GestaltMatcher-Arc Embedding Analysis}
\begin{center}

\captionof{table}{
Disease-specific changes in GestaltMatcher-Arc representation geometry from
Baseline to PD50. Compactness is mean cosine distance to the class centroid
(lower is better); separation is cosine distance to the nearest foreign-class
centroid (higher is better). Sep./Comp. is their ratio. Changes are relative
percentages.
}
\label{tab:embedding_results}

\vspace{2pt}

\scriptsize
\setlength{\tabcolsep}{2.2pt}
\renewcommand{\arraystretch}{1.05}

\resizebox{\columnwidth}{!}{%
\begin{tabular}{lrrrr}
\toprule
\textbf{Disease} &
\textbf{Train $N$} &
\textbf{$\Delta$ Comp.} &
\textbf{$\Delta$ Sep.} &
\textbf{$\Delta$ Sep./Comp.} \\
\midrule

Cornelia de Lange      & 312 & \cellcolor{bestresult}-9.35  & \cellcolor{worstresult}-9.86 & \cellcolor{worstresult}-0.56 \\
Williams--Beuren       & 191 & \cellcolor{bestresult}-2.33  & \cellcolor{bestresult}9.51   & \cellcolor{bestresult}12.13 \\
Noonan                  & 145 & \cellcolor{bestresult}-9.06  & \cellcolor{bestresult}2.07   & \cellcolor{bestresult}12.24 \\
Kabuki                  & 144 & \cellcolor{bestresult}-9.25  & \cellcolor{bestresult}4.76   & \cellcolor{bestresult}15.44 \\
KBG                     & 125 & \cellcolor{bestresult}-13.54 & \cellcolor{bestresult}18.56  & \cellcolor{bestresult}37.12 \\
Angelman                & 111 & \cellcolor{bestresult}-3.32  & \cellcolor{bestresult}18.46  & \cellcolor{bestresult}22.53 \\
Rubinstein--Taybi       & 81  & \cellcolor{bestresult}-3.36  & \cellcolor{bestresult}4.91   & \cellcolor{bestresult}8.55 \\
Smith--Magenis          & 71  & \cellcolor{worstresult}2.35  & \cellcolor{bestresult}23.91  & \cellcolor{bestresult}21.07 \\
Nicolaides--Baraitser   & 68  & \cellcolor{bestresult}-5.69  & \cellcolor{bestresult}9.51   & \cellcolor{bestresult}16.12 \\
22q11.2 deletion        & 41  & \cellcolor{worstresult}0.94  & \cellcolor{bestresult}23.91  & \cellcolor{bestresult}22.76 \\

\bottomrule
\end{tabular}%
}

\end{center}

\fi
\end{document}